\documentclass[conference]{IEEEtran}
\IEEEoverridecommandlockouts
\usepackage{amsmath,amsfonts}
\usepackage{algorithmic}
\usepackage{algorithm}
\usepackage{array}
\usepackage[caption=false,font=normalsize,labelfont=sf,textfont=sf]{subfig}
\usepackage{textcomp}
\usepackage{stfloats}
\usepackage{url}
\usepackage{verbatim}
\usepackage{graphicx}
\usepackage{cite}
\usepackage{url}
\usepackage{booktabs}
\usepackage{multirow}
\usepackage{amsmath}
\usepackage{pifont}

\def\BibTeX{{\rm B\kern-.05em{\sc i\kern-.025em b}\kern-.08em
    T\kern-.1667em\lower.7ex\hbox{E}\kern-.125emX}}

\begin{document}

\title{Quaternion Wavelet-Conditioned Diffusion Models for Image Super-Resolution
\thanks{This work was partly supported by ``Ricerca e innovazione nel Lazio - incentivi per i dottorati di innovazione per le imprese e per la PA - L.R. 13/2008" of Regione Lazio, under grant number 21027NP000000136, and by the European Union under the Italian National Recovery and Resilience Plan (NRRP) of NextGenerationEU, ``Rome Technopole" (CUP B83C22002820006)—Flagship Project 5: ``Digital Transition through AESA radar technology, quantum cryptography and quantum communications", and under the partnership on ``Future Artificial Intelligence Research” (PE0000013 - FAIR - Spoke 5: High Quality AI).}
}

\author{\IEEEauthorblockN{Luigi Sigillo, Christian Bianchi, Aurelio Uncini, and Danilo Comminiello}
        \IEEEauthorblockN{\textit{Dept. Information Engineering, Electronics and Telecommunications (DIET), Sapienza University of Rome, Italy}}
Email: luigi.sigillo@uniroma1.it.
}




\maketitle

\begin{abstract}
Image Super-Resolution is a fundamental problem in computer vision with broad applications spacing from medical imaging to satellite analysis. The ability to reconstruct high-resolution images from low-resolution inputs is crucial for enhancing downstream tasks such as object detection and segmentation. While deep learning has significantly advanced SR, achieving high-quality reconstructions with fine-grained details and realistic textures remains challenging, particularly at high upscaling factors. Recent approaches leveraging diffusion models have demonstrated promising results, yet they often struggle to balance perceptual quality with structural fidelity.
In this work, we introduce ResQu a novel SR framework that integrates a quaternion wavelet preprocessing framework with latent diffusion models, incorporating a new quaternion wavelet- and time-aware encoder. Unlike prior methods that simply apply wavelet transforms within diffusion models, our approach enhances the conditioning process by exploiting quaternion wavelet embeddings, which are dynamically integrated at different stages of denoising. Furthermore, we also leverage the generative priors of foundation models such as Stable Diffusion.
Extensive experiments on domain-specific datasets demonstrate that our method achieves outstanding SR results, outperforming in many cases existing approaches in perceptual quality and standard evaluation metrics.
The code is available at \url{https://www.github.com/Fascetta/ResQu}.
\end{abstract}


\section{Introduction}

\begin{figure}[!t]
\centering
\includegraphics[width=\linewidth]{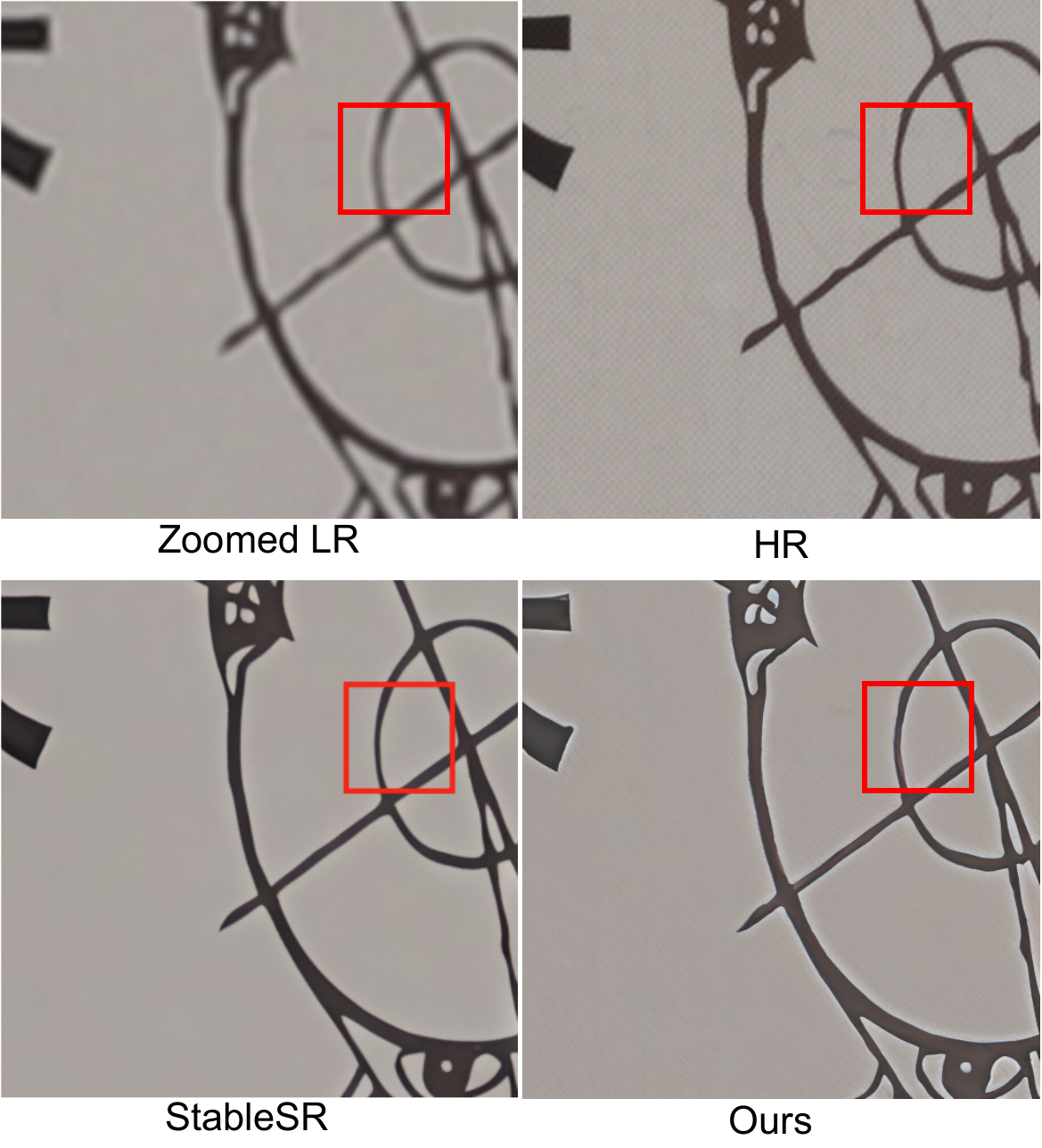}
\caption{Qualitative comparisons on real-world images, upscaled from 128 to 512 pixels, we compare StableSR \cite{wang2023exploiting}, and the ground truth with our results displayed in the final column. The visual evaluations underscore differences in image quality, clarity, and the enhancement of fine details achieved by our model during the super-resolution process. Noticeable to see that the background on StableSR results is plain while in the ground truth it is coarse, like in the output of our model.}
\label{fig:sample_all}
\end{figure}

Image Super-Resolution (SR) is a cornerstone problem in computer vision, with far-reaching implications for applications ranging from medical imaging to satellite analysis and beyond \cite{lepcha2023image,sigilloMWT_2025}. The ability to reconstruct high-resolution (HR) images from their low-resolution (LR) counterparts is not only a technical challenge but also a practical necessity. Indeed, LR images often hinder the performance of downstream tasks such as object detection, segmentation, and classification \cite{kim2024beyond}. Despite decades of research, achieving SR that preserves fine-grained details and realistic textures, especially at high upscaling factors, remains an open problem.

Traditional SR methods, such as interpolation, frequency-domain transformations, and filtering-based techniques, have provided foundational solutions but often fall short of generating perceptually convincing results. These approaches typically struggle to recover high-frequency details, leading to artifacts and blurring that degrade image quality \cite{10183382}.
The field of super-resolution (SR) has undergone a transformative shift with the advent of deep learning. Convolutional neural networks (CNNs), Vision Transformers (ViT), and more recently, generative models have all set new benchmarks for SR performance. 
Among these advancements, diffusion models have emerged as a particularly promising paradigm for image generation. Studies have demonstrated their unparalleled capabilities in generating coherent images by iteratively refining noise into structured outputs \cite{Ruiz_2023_CVPR, lopez2025guessithinkstreamlined, Takagi_2023_CVPR, wang2024lavie}.
Building upon this progress, there have been many recent advances in the application of diffusion models to super-resolution tasks \cite{saharia2021image, shang2024resdiff, gao2023implicit, 10650042, wu2024seesr}. 

Contemporary research in diffusion-based super-resolution has increasingly focused on exploring the synergistic potential of integrating wavelet-based techniques, leveraging their complementary strengths in multi-resolution analysis and feature decomposition \cite{huang2024wavedm}. For instance, \cite{phung2023wavelet} introduced a wavelet-based diffusion framework that bridges the speed gap in generative models by adaptively handling low- and high-frequency components through wavelet decomposition. Building on this, other works \cite{aloisi2024waveletdiffusionganimage, 10651227} have further explored wavelet diffusion approaches tailored specifically for SR tasks, establishing a strong foundation for integrating wavelet transforms with generative models. These works highlight the importance of multi-scale feature extraction and adaptive processing in achieving state-of-the-art SR performance.

Motivated by these recent progresses in the field, we propose a novel SR framework that integrates a quaternion wavelet preprocessing framework \cite{sigillo2024generalizingmedicalimagerepresentations} with latent diffusion models. Unlike prior works that simply combine wavelet transforms with diffusion models, our approach introduces a quaternion wavelet- and time-aware encoder that significantly enhances the conditioning process during the denoising steps. This encoder exploits quaternion wavelet embeddings, which are dynamically integrated at different stages of the denoising process, enabling fine-grained control over the generation of SR images. By leveraging the latent representations learned by state-of-the-art text-to-image generative models such as Stable Diffusion (SD) \cite{Rombach2021HighResolutionIS}, our framework not only advances existing SR methodologies but also establishes a novel architectural paradigm that synthesizes generative priors with super-resolution techniques.

The quaternion wavelet embeddings capture both low-frequency approximations and high-frequency details, which are weighted and conditioned at multiple scales during denoising by our encoder. This multi-scale conditioning mechanism, combined with the generative priors of SD, allows our framework to achieve superior performance in quantitative metrics, and also competitive results in qualitative assessments. To validate our approach, we conduct extensive experiments with additional ablation studies on diverse datasets, demonstrating the robustness and versatility of our method in real-world scenarios.

The primary contributions of this work are as follows:
\begin{enumerate}
    \item We introduce a latent diffusion model for image SR that integrates quaternion wavelet features, and generative priors.
    \item  We propose a novel quaternion wavelet- and time-aware encoder that enhances the conditioning process at multiple scales during denoising, significantly improving the performance of pre-trained foundation models like SD.
    \item We demonstrate through extensive experiments that our approach outperforms existing methods in both perceptual quality and quantitative metrics.
    \item We perform ablation studies on architecture design and also on domain-specific datasets.
\end{enumerate}
This paper is organized as follows. In Section~\ref{sec:background}, we provide the theoretical background and discuss the challenges and applications of SR. In Section~\ref{sec:method} we introduce our proposed method, StableShip-SR, detailing its architecture and key innovations. In Section~\ref{sec:experiment} we present our experimental results and comparisons with state-of-the-art methods. Finally, Section~\ref{sec:conclusion} concludes the paper with a discussion of our findings and potential directions for future research.

\section{Background}
\label{sec:background}

Image SR is a fundamental task in computer vision, aimed at restoring high-resolution (HR) images from degraded low-resolution (LR) observations. Given an LR image \(\tilde{\mathbf{x}}\), the goal is to reconstruct an HR image \(\mathbf{x}\), where the relationship between the two is modeled as:  
\begin{equation}
\tilde{\mathbf{x}} = (\mathbf{x} \otimes k) + n,    
\end{equation}
where \(k\) represents the degradation matrix, \(\otimes\) denotes a convolution-like operation, and \(n\) is a noise term. This task is inherently ill-posed, as multiple HR solutions can correspond to the same LR input, making reconstruction highly ambiguous and challenging \cite{dai2019second,dong2014learning, dong2015image, dong2016accelerating}.  

\textbf{Deep Learning-Based Approaches.} Early approaches to SR mainly relied on traditional methods such as bicubic interpolation and frequency domain filtering, which assumed simple degradation models and achieved limited success in recovering fine details. The advent of deep learning marked a turning point, with models like SRCNN \cite{dong2014learning} pioneering the application of CNNs to SR tasks. Subsequent architectures, including EDSR \cite{ledig2017photorealistic}, ESRGAN \cite{Wang2018ESRGANES}, and RCAN \cite{Zhang2018ImageSU}, demonstrated the ability of CNNs to learn complex mappings between LR and HR domains, significantly improving the perceptual quality of SR images.

CNNs dominated the early landscape of deep learning-based SR research until the introduction of Vision Transformers (ViT) \cite{dosovitskiy2021image} and SwinIR \cite{swinir}, the latter of which builds upon ViT by leveraging shifted window attention mechanisms to efficiently process large-scale images and model long-range dependencies. Despite these advancements, such models often exhibit limitations in comprehensively capturing global semantic context, a critical shortcoming in complex real-world applications.

\textbf{Generative Models for SR.} Recent advancements have seen a paradigm shift toward generative models, including GANs and diffusion models, for SR tasks. GAN-based methods, such as PULSE \cite{Menon2020PULSESP}, utilize adversarial training to generate perceptually realistic details. However, the instability of GAN training and the tendency to produce unnatural artifacts remain notable challenges. Blind SR approaches, such as BSRGAN \cite{zhang2021designing} and Real-ESRGAN \cite{wang2021real}, introduce sophisticated degradation pipelines that more accurately reflect real-world conditions. Despite these advances, many SR methods still rely on predefined degradation assumptions, limiting their generalizability.

Diffusion models \cite{Song2020DenoisingDI}, offer a more stable and controllable framework for image generation. These models iteratively refine noisy inputs, producing HR images by leveraging the data distribution during a denoising process.  
Among diffusion-based SR models, SR3 \cite{saharia2021image} has demonstrated remarkable performance, employing a conditional diffusion process guided by a modified U-Net with G-Blocks sourced from BigGAN \cite{brock2019large}. However, SR3’s reliance on pixel-space diffusion makes it computationally expensive. Latent diffusion models (LDMs) \cite{Rombach2021HighResolutionIS} address these challenges by shifting the denoising process to a compressed latent space, significantly reducing computational overhead while maintaining high-resolution outputs.  
StableSR \cite{wang2023exploiting} fine-tunes Stable Diffusion (SD) \cite{Rombach2021HighResolutionIS}, a pre-trained text-to-image latent diffusion model, by introducing time-aware encoders to balance fidelity and perceptual quality. Similarly, DiffBIR \cite{lin2025diffbir} employs a two-stage process that first reconstructs the image and then enhances details using the diffusion prior. These methods effectively leverage the extensive generative prior encapsulated in T2I models, trained on vast and diverse datasets, to address complex real-world SR challenges.  

\textbf{Our approach.} Despite their successes, diffusion-based SR methods face certain limitations, including high computational costs and difficulties in preserving domain-specific details. In this work, we introduce a novel approach that combines latent diffusion models with quaternion wavelets exploiting QUAVE \cite{sigillo2024generalizingmedicalimagerepresentations}, a preprocessing framework originally designed for medical imaging tasks such as segmentation and reconstruction. QUAVE utilizes quaternion wavelet transforms to decompose images into low- and high-frequency components, capturing a rich multidimensional representation of the data.  

Our method integrates QUAVE as a conditioning mechanism in the latent diffusion process, enabling the preservation of fine-grained details critical for domain-specific tasks. Using the spatial and frequency domain features extracted by QUAVE, our approach provides the SR model with richer input, improving its ability to generalize across diverse datasets. Additionally, unlike methods that require training from scratch, our approach fine-tunes pre-trained SD \cite{Rombach2021HighResolutionIS}, significantly reducing computational requirements while maintaining state-of-the-art performance.

\section{The Proposed ResQu Method}
\label{sec:method}
\begin{figure}[!t]
    \centering
    \includegraphics[width=0.95\linewidth]{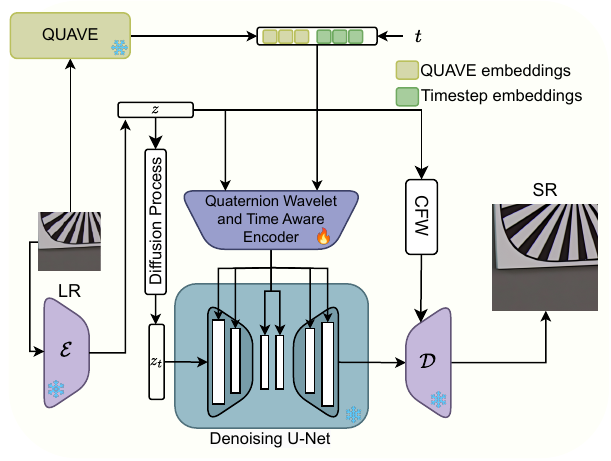}
    \caption{Overview of ResQu Super-Resolution framework. We pre-trained QUAVE and trained only our encoder, the quaternion embeddings- and time- aware encoder. The other parts of the framework are frozen, which speeds up the overall training process.}
    \label{fig:architecture}
\end{figure}

In this section, we describe our proposed method, which leverages the generative prior of SD \cite{Rombach2021HighResolutionIS} and introduces a novel quaternion wavelet and time-aware encoder for image super-resolution. By integrating quaternion wavelet embeddings and temporal conditioning into the diffusion process, our approach achieves high-fidelity reconstructions while preserving domain-specific details.



\subsection{Quaternion Wavelet and Time-Aware Encoder}
\textbf{Real Wavelet Transform.} The one-dimensional discrete wavelet transform (1D-DWT) characterizes a signal $f(t)$ using scaling function $\phi(t)$ and wavelet function $\psi(t)$ \cite{barford1992introduction}. The 2D-DWT extends this through tensor products along spatial dimensions, yielding scaling function $\phi(x)\phi(y)$ and three wavelets $\psi(x)\psi(y)$, $\phi(x)\psi(y)$, and $\psi(x)\phi(y)$ for diagonal, horizontal, and vertical features \cite{vetterli1992wavelets, chan2004quaternion}. This decomposition produces a low-frequency (LL) and three high-frequency (LH, HL, HH) sub-bands. However, DWT lacks translation invariance and phase information preservation, making wavelet coefficients sensitive to acquisition variations \cite{selesnick2005dual, chan2008coherent}.

\textbf{Quaternion Wavelet Transform.} The Quaternion Wavelet Transform (QWT) integrates four quaternion wavelet transforms, constructed through the combination of a real DWT and its three corresponding Hilbert transforms along the $x$, $y$, and $xy$ axes. This transformation decomposes the input image into four quaternion wavelet sub-bands, yielding a total of 16 real sub-bands.
The QWT decomposition follows a structure analogous to the DWT, producing a scaling function denoted as $\phi_q$ and three directional wavelets: $\psi^D_q$, $\psi^V_q$, and $\psi^H_q$, corresponding to diagonal, vertical, and horizontal orientations, respectively \cite{zhancheng2020medical, 4526699}.

\textbf{Quaternion Wavelets Embeddings.} 
Directly utilizing all sixteen QWT sub-bands as input to a neural model introduces redundancy and inefficiency \cite{grouse}. To address these challenges, we employ QUAVE \cite{sigillo2024generalizingmedicalimagerepresentations}, a learning-based approach that extracts only the features from the most informative sub-bands of the QWT, which utilizes the dual-tree quaternion wavelet transform (QWT) \cite{Chan2004QWT}.

Our quaternion wavelet, and time aware encoder leverages quaternion wavelet embeddings, obtained from QUAVE, to provide multidimensional representations that capture low- and high-frequency information essential for super-resolution.
Our encoder processes LR input images to extract the enhanced features. These features, enriched by information from both the frequency and spatial domains, are injected into the denoising process via spatial feature transformations (SFT) \cite{8578168}.

During the diffusion process, the QUAVE embeddings together within the timestep embeddings are passed through the encoder and used to modulate the intermediate feature maps of the U-Net via SFT. This ensures that the model incorporates both global and local details, enhancing the fidelity and realism of the reconstructed HR image.  

\textbf{Time-Aware Embeddings.} In addition to quaternion wavelet embeddings, the encoder incorporates temporal information by embedding the timestep \(t\) into the conditioning mechanism. Early in the denoising process, when the latent representation has low SNR and contains significant noise, the encoder provides strong guidance using the enhanced features derived from QUAVE. This ensures that the structural integrity of the image is preserved during the initial stages of generation. 
This is because, as discovered in \cite{9879163}, when the SNR increases and the latent representation becomes more refined, the encoder reduces its influence, allowing the diffusion model to focus on fine-grained details. This adaptive behavior is crucial for achieving a balance between preserving global structure and enhancing local textures. The known noise schedule of the diffusion process enables the encoder to adjust its conditioning strength dynamically at each timestep, ensuring optimal guidance throughout the super-resolution pipeline.  

\subsection{Latent Diffusion Process}
The conditioning process is defined by concatenating the QUAVE and timestep embedding tensors into a unified vector \( b \in \mathbb{R}^{1024} \), as depicted in Fig. \ref{fig:architecture}. This vector is processed alongside the latent representation \( z\in \mathbb{R}^{h\times w\times c} \) of the input low-resolution image \( x\in \mathbb{R}^{3\times H\times W} \), where \( z=\mathcal{E}(x) \) is obtained via the encoder \( \mathcal{E} \).  

For our ResQu we leverage the pre-trained variational autoencoder (VAE) from \cite{Rombach2021HighResolutionIS}, with both the encoder \( \mathcal{E} \) and decoder \( \mathcal{D} \) kept frozen. Additionally, we employ their time-conditional U-Net backbone for the latent diffusion process. 

To enhance the conditioning mechanism, we introduce the encoder \( \delta_\theta \), designed to mirror the U-Net encoder architecture. This structure enables the extraction of conditioning embeddings at multiple scales, which are subsequently incorporated into the conditional denoising autoencoder \( \epsilon_\theta \), facilitating a more refined and hierarchical feature representation.  
We optimize the following objective:  
\begin{equation}
L = \mathbb{E}_{\mathcal{E}(x), y, \epsilon \sim \mathcal{N}(0,1), t,c}\left[\|\epsilon - \epsilon_\theta(z_t, \delta_\theta(c, t, z))\|_2^2\right].    
\end{equation}



By integrating quaternion wavelet embeddings and temporal guidance, our method effectively leverages the generative prior of SD for ship image super-resolution. This combination ensures high-fidelity reconstructions, enhanced detail preservation, and robustness across diverse scenarios.

\begin{table*}
\caption{Quantitative comparison with state-of-the-art methods on both synthetic and real-world benchmarks. The best performance for each metric is in \textbf{bold}, while the second-best is \underline{underlined}.}
\label{tab:objective_results_general}
\begin{tabular}{l|c|cccccccc}
\toprule
 & \textbf{Metrics} & RealSR \cite{ji2020real} & BSRGAN  \cite{zhang2021designing}& FeMaSR \cite{chen2022real} & R-ESRGAN+ \cite{wang2021real} & ResShift \cite{yue2024resshift} & LDM \cite{Rombach2021HighResolutionIS} & StableSR \cite{wang2023exploiting} & \textbf{Ours} \\
\midrule
\multirow{4}{*}{\rotatebox{90}{\scriptsize{\textbf{DIV2K-Val}}}}
& PSNR $\uparrow$ & \underline{24.62} & 24.58 & 22.97 & 24.29 & 24.53 & 20.58 & 23.26 & \textbf{25.21} \\
& SSIM $\uparrow$ & 0.5970 & 0.6269 & 0.5887 & \underline{0.6372} & 0.7323 & 0.5762  & 0.5726  & \textbf{0.645}\\
& LPIPS $\downarrow$ & 0.5276 & 0.3351 & 0.3126 & \textbf{0.3112} & 0.4406 & 0.3199  & \underline{0.3114} & 0.4161 \\
& FID $\downarrow$ & 49.49 & 44.22 & 35.87 & 37.64 & 49.16 & 26.47 & \textbf{24.44} & \underline{25.43}\\
\midrule
\multirow{3}{*}{\rotatebox{90}{\scriptsize{\textbf{RealSR}}}}
& PSNR $\uparrow$ & \underline{25.56} & 24.70 & 23.58 & 24.33 & 24.79 & 22.26  & 23.55 & \textbf{26.45}\\
& SSIM $\uparrow$ & 0.7390 & \textbf{0.7651} & 0.7132 & 0.7456 & 0.7423 & 0.6462 & 0.7080 & \underline{0.7627}\\
& LPIPS $\downarrow$ & 0.3570 & \underline{0.2713} & 0.2937 & \textbf{0.2524} & 0.3134 & 0.3159  & 0.3002 & 0.3215 \\
\midrule
\multirow{2}{*}{\rotatebox{90}{\scriptsize{\textbf{DRealSR}}}}
& PSNR $\uparrow$ & 27.79 & 26.18 & 24.56 & 25.82 & \underline{27.87} & 23.39 & 24.85 & \textbf{29.69}\\
& SSIM $\uparrow$ & \underline{0.8148} & 0.8028 & 0.7569 & 0.8052 & 0.8056 & 0.7448 & 0.7536 & \textbf{0.8211}\\
& LPIPS $\downarrow$ & 0.3938 & \textbf{0.2929} & 0.3157 & \underline{0.2818} & 0.5408 & 0.3379 & 0.3284 & 0.3332 \\
\bottomrule
\end{tabular}
\end{table*}

\section{Experimental Results}
\label{sec:experiment}
\begin{figure*}[!t]
\centering
\includegraphics[width=\textwidth]{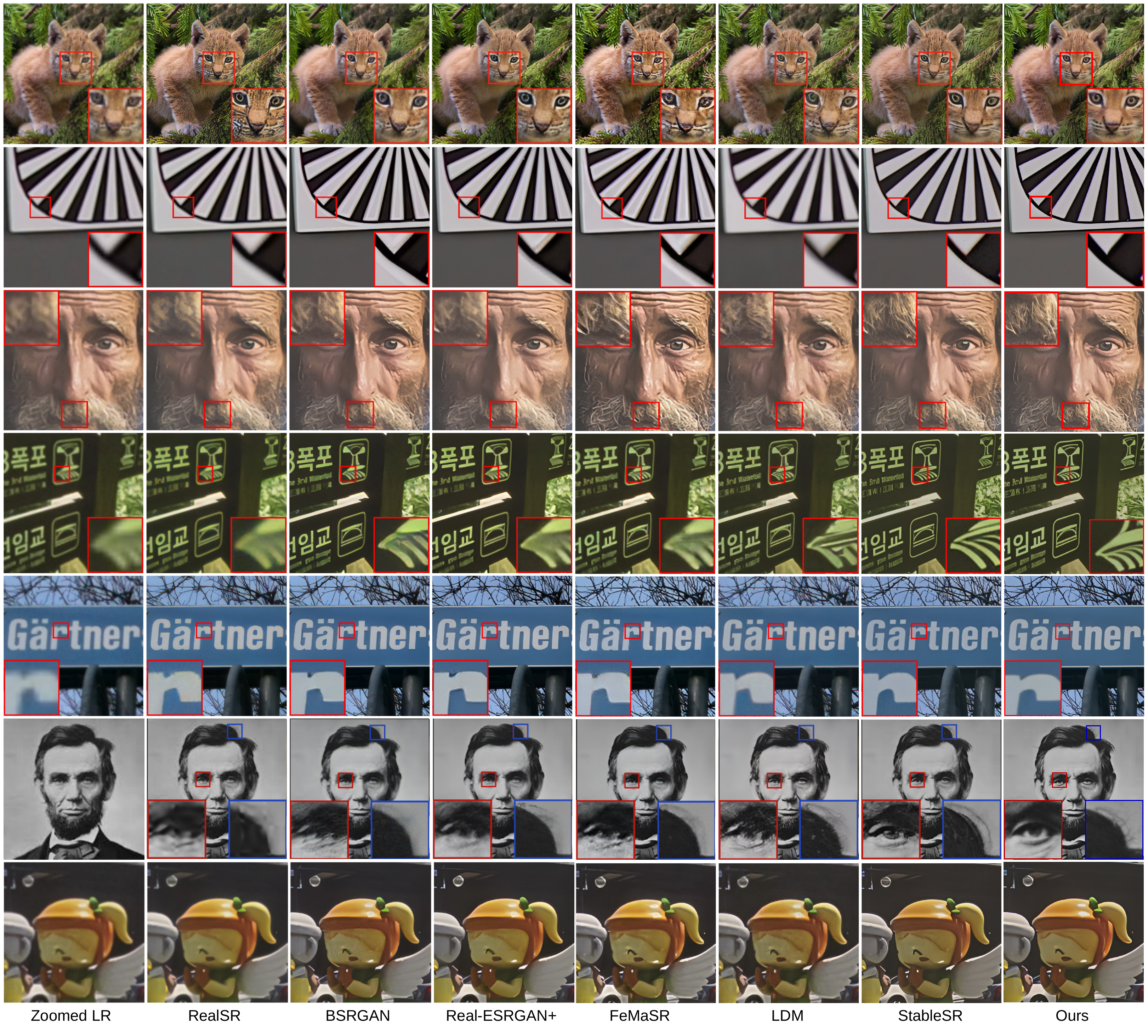}
\caption{Comparison of the LR input image with SR outputs generated by state-of-the-art methods and our proposed model on the DRealSR \cite{wei2020component} and RealSR \cite{cai2019toward} datasets. The red bounding box highlights a zoomed-in region, illustrating the superior resolution and detail preservation achieved by our method compared to existing approaches.}
\label{fig:zoomed}
\end{figure*}
\subsection{Experimental Setup}
All experiments were conducted using an NVIDIA RTX A6000 GPU with 48GB of VRAM. We employ the AdamW optimizer with a learning rate of 5e-05. The batch size was set limited to $6$ to balance memory efficiency and stable convergence. The model was trained for 14k iterations, starting from the released checkpoints of StableSR \cite{wang2023exploiting}. We perform super-resolution going from a resolution of 128x128 to 512x512, thus with a x$4$ factor. QUAVE was pre-trained separately for 45k steps on a combined dataset consisting of DIV2K \cite{agustsson2017ntire}, Flickr2K \cite{timofte2017ntire}, and OutdoorSceneTraining (OST) \cite{wang2018recovering}, ensuring robust feature extraction across diverse image domains.

\subsection{Quantitative Evaluation}
We evaluated our model's performance across three benchmark datasets: DRealSR \cite{wei2020component}, RealSR \cite{cai2019toward}, and DIV2K, each posing unique challenges for super-resolution. The quantitative results, summarized in Table \ref{tab:objective_results_general}, demonstrate the superior performance of our approach.
We assess our super-resolution model using PSNR and SSIM (calculated on the Y channel in YCbCr space), FID, and LPIPS, ensuring a comprehensive pixel-level accuracy and perceptual quality evaluation.
PSNR quantifies reconstruction fidelity by measuring pixel-wise differences, but it fails to align with human perception. SSIM improves upon PSNR by incorporating structural similarity, yet it remains limited in capturing perceptual realism.
Conversely, FID evaluates realism by comparing feature distributions in a learned embedding space, effectively capturing perceptual discrepancies. LPIPS, leveraging deep network activations, provides a more human-aligned similarity measure. These perceptual metrics better reflect image quality, particularly in adversarial and generative models, where traditional pixel-wise approaches fall short \cite{saharia2021image}.

On the DIV2K \cite{agustsson2017ntire} dataset, our model exhibited particularly strong performance, achieving a higher SSIM and PSNR while having a competitive FID and LPIPS score. These results indicate the effectiveness of our method in maintaining both structural fidelity and perceptual quality, even on high-resolution images with diverse textures.

For the RealSR \cite{cai2019toward} and DRealSR \cite{wei2020component} datasets, which include real-world degradations, our approach demonstrated strong generalization capabilities, achieving higher PSNR and SSIM which also indicates that our model effectively preserves fine-grained textures, leading to perceptually superior reconstructions. 

\subsection{Qualitative Analysis}
To further evaluate the effectiveness of our approach, we conduct a qualitative comparison of super-resolved images, presented in Figure \ref{fig:zoomed}. The visual results underscore several key advantages of our method over existing techniques.

Our model excels in reconstructing intricate structural details and fine textures. The quaternion wavelet-based conditioning effectively preserves sharp edges while mitigating artifacts commonly observed in baseline methods. This is particularly evident in the Lincoln portrait (the second to last image in the grid in Fig. \ref{fig:zoomed}), where our approach retains subtle facial features, such as fine wrinkles and hair textures, with minimal over smoothing. Compared to alternative methods, our model produces a more naturalistic representation, avoiding excessive blurring or unnatural sharpness.

Additionally, our method demonstrates enhanced robustness in reconstructing complex high-frequency textures, such as those present in the lynx's fur (the first row in the grid in Fig. \ref{fig:zoomed}). While traditional SR methods often introduce noise or fail to recover the natural stochastic patterns in such textures, our approach maintains structural coherence, effectively capturing both coarse and fine-scale fur patterns. The reduction in visual artifacts and improved textural realism contribute to a more perceptually pleasing output.

A particularly challenging aspect of super-resolution lies in handling textual elements, as fine details must be recovered with high fidelity to maintain readability. Our approach significantly outperforms previous methods in this regard, as demonstrated in the writing of "Gartner" text example (the fifth row in the grid of Fig. \ref{fig:zoomed}), where letter edges remain crisp, and subtle font characteristics are preserved without aliasing or distortion. Traditional methods tend to introduce blurring or edge artifacts, whereas our model retains sharpness and legibility, highlighting its advantages in text-based image enhancement applications.

Overall, our experimental findings validate the effectiveness of the proposed method in achieving state-of-the-art super-resolution performance. The ability to balance structural fidelity and perceptual quality across diverse content types—ranging from human portraits and natural textures to text elements—demonstrates its broad applicability in real-world image enhancement tasks.


\subsection{Ablation Studies}
\label{sec:ablation}
\begin{figure}[!t]
    \centering
    \includegraphics[width=0.8\linewidth]{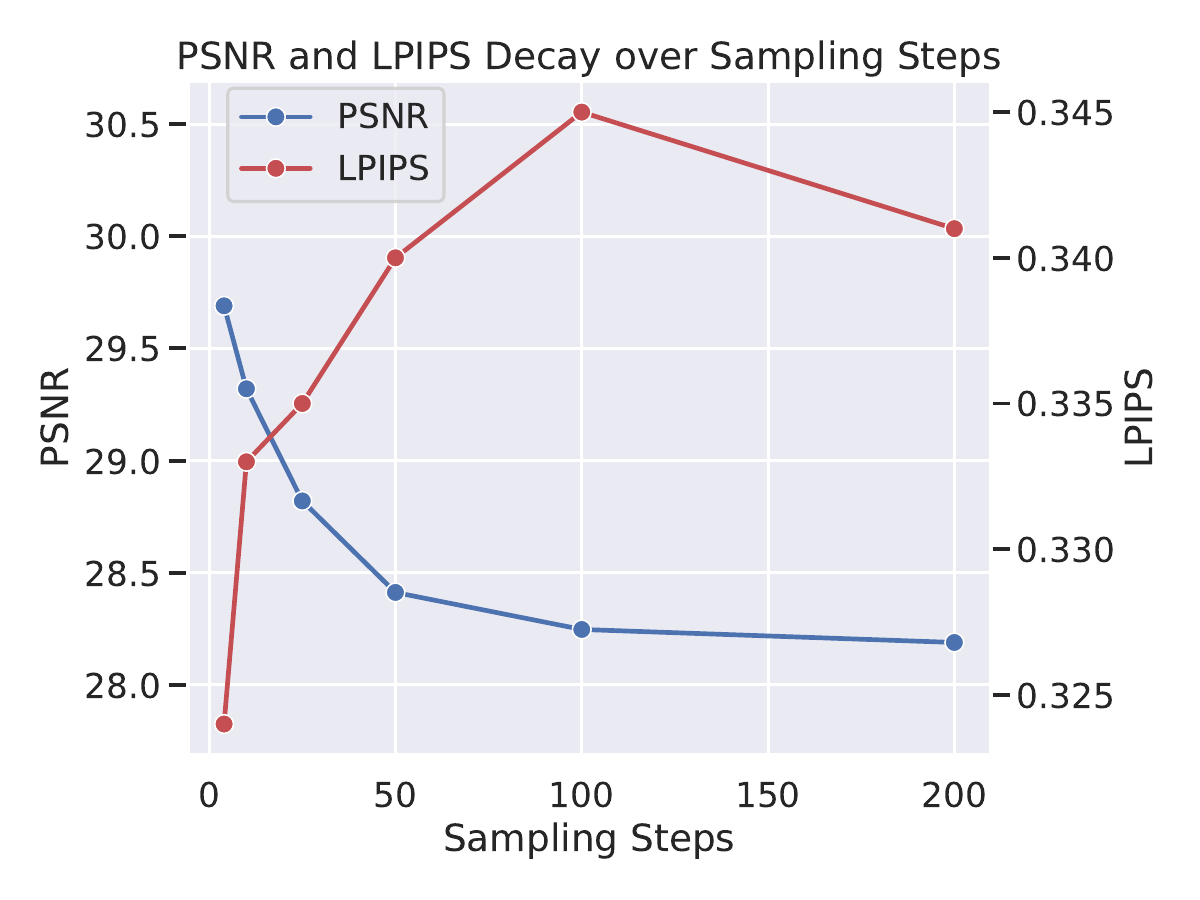}
    \caption{Impact of the number of sampling steps on key evaluation metrics. Reducing the steps improves efficiency while maintaining performance, highlighting the inherent speed advantage of our model.}
    \label{fig:ablation_steps}
\end{figure}

\textbf{Ablation Study on the number of sampling steps.} 
For a fair comparison, we set the number of sampling steps to $200$, following the configuration of StableSR \cite{wang2023exploiting}. Subsequently, we explored reducing this number and observed that our model benefits from fewer sampling steps. Figure \ref{fig:ablation_steps} illustrates the impact of this reduction on key evaluation metrics.
We discover that reducing the number of sampling steps introduces a tradeoff among the evaluation metrics. Specifically, we observe an increase in PSNR and SSIM, which suggests improved pixel-wise and structural similarity to the ground truth. However, LPIPS also increases, indicating a decline in perceptual quality. This suggests that while fewer steps enhance fidelity in terms of traditional similarity metrics, they may compromise perceptual realism. Therefore, the choice of sampling steps depends on the desired balance between pixel-wise accuracy and perceptual quality in practical applications.

\textbf{Ablation Study on the Controllable Feature Wrapping.} To thoroughly evaluate the impact of different Controllable Feature Wrapping (CFW) implementations, we conduct a comparative analysis between our custom-trained CFW module and the one introduced by StableSR \cite{wang2023exploiting}. 
The training pipeline generates synthetic LR-HR pairs specifically designed to enhance the ability of the module to maintain realistic textures and natural image statistics.

Table \ref{tab:cfw_results} presents quantitative results across different metrics for two model variants: our model with custom-trained CFW and our model utilizing CFW module pretrained by StableSR. 
The improvements are particularly notable in areas with complex geometrical patterns and sharp edges, suggesting that our training approach effectively leverages the structural information captured by the quaternion wavelet features, this is confirmed by the SSIM and PSNR metrics. 

\begin{figure}[!t]
    \centering
    \includegraphics[width=0.9\linewidth]{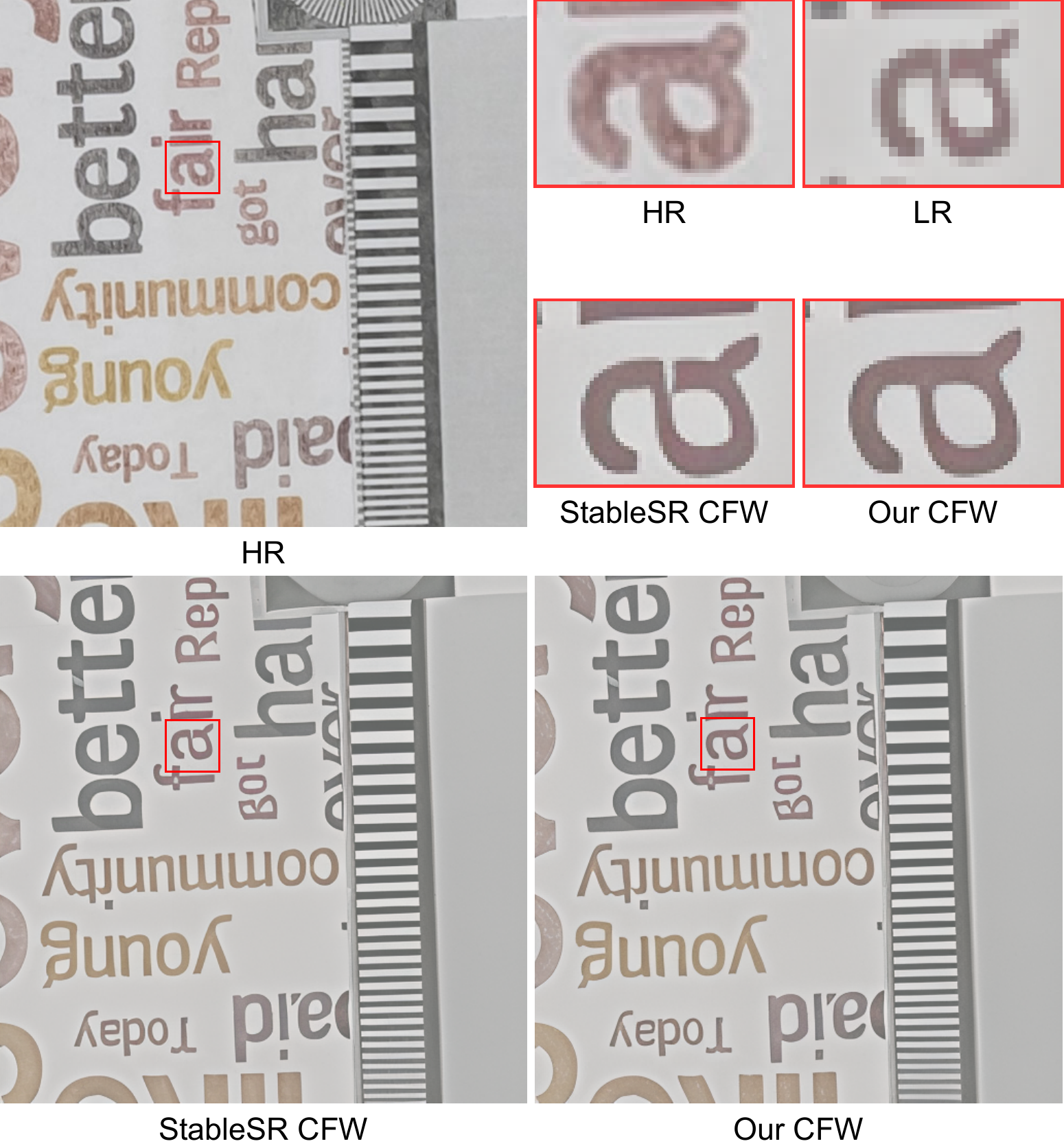}
    \caption{Visual comparison of diverse super-resolution results generated by changing the CFW. The images demonstrate the impact of our CFW enhanced by our encoder, producing more structured details.}
    \label{fig:visual_ablation_steps}
\end{figure}

These results suggest that the choice of CFW implementation should be guided by specific application requirements. When structural fidelity is paramount, our custom-trained CFW provides superior results. In Figure \ref{fig:visual_ablation_steps} there is a visual comparison of the output generated with the different CFW.

\begin{table}[t] 
\centering 
\caption{Metrics obtained with different versions of the CFW with our ResQu.}
\label{tab:cfw_results}
\begin{tabular}{lcccccc} 
\toprule
\textbf{Dataset} & \textbf{CFW} & \textbf{PSNR} \textuparrow & \textbf{SSIM} \textuparrow & \textbf{FID} \textdownarrow \\
\midrule
\multirow{2}{*}{DIV2K} & StableSR & \textbf{$23.34$} & $0.562$ & $\textbf{24.52}$\\
 & Ours & $\textbf{23.61}$ & $\textbf{0.591}$ &  $25.42$ \\
 \midrule
\multirow{2}{*}{RealSR} & StableSR & $23.26$ & $0.702$ & $\textbf{129}$\\ 
 & Ours & $\textbf{24.98}$ & $\textbf{0.714}$ & $134$\\
 \midrule
\multirow{2}{*}{DrealSR} & StableSR & $27.94$ & $0.734$ & $151$\\ 
 & Ours & $\textbf{28.18}$ & $\textbf{0.755}$ & $\textbf{151}$\\
\bottomrule
\end{tabular}
\end{table}

\textbf{Ablation Study on ShipSpotting Dataset.} To assess the generalization capability of our model and its applicability to domain-specific tasks without retraining, we conduct a zero-shot evaluation on the ShipSpotting dataset \cite{10650042}. This dataset comprises diverse maritime vessel images captured under varying conditions, presenting unique super-resolution challenges due to complex textures, fine structural details, and significant variations in scale.

As shown in Table \ref{tab:objective_results_ships}, despite not being explicitly trained on this dataset, our model achieves performance comparable to fully trained approaches. Notably, ResQu attains SSIM and PSNR scores that closely match those of specialized models while maintaining a competitive FID score. This demonstrates its ability to generalize effectively to unseen domains, highlighting its robustness in handling domain-specific scenarios without requiring retraining.

The improvements are particularly evident in the reconstruction of intricate maritime structures, such as ship rigging, antenna arrays, and deck equipment, where traditional methods often struggle. Our quaternion-based representation captures essential phase relationships, ensuring sharper edges and enhanced texture preservation, which are critical for maritime imagery. Additionally, the multi-scale conditioning mechanism enables our model to consistently recover both large-scale structural elements (e.g., hull details) and fine-grained features (e.g., navigation equipment and water surface textures).

These findings validate the versatility of our approach, demonstrating that it not only achieves state-of-the-art results on its primary benchmark datasets but also extends effectively to domain-specific applications in a zero-shot setting.
\begin{table}[t]
\centering
\caption{FID results obtained on the Shipspotting \cite{10650042} dataset.}
\label{tab:objective_results_ships}
\begin{tabular}{cc|ccc}
\toprule
 & Model  & SSIM $\uparrow$ & PSNR $\uparrow$ & FID $\downarrow$  \\
\midrule  
\multirow{3}{*}{Full Training}
& SR3\cite{saharia2021image}  & 0.569 & 21.07  & 16.59  \\ 
& StableSR\cite{wang2023exploiting}  & 0.554 & 20.29  &   \underline{12.41}  \\
& StableShip-SR \cite{10650042} & 0.561 & 20.51 &  \textbf{11.72} \\ 
\midrule
Zero-Shot & ResQu  & \textbf{0.611} & \textbf{22.01} &  14.63 \\ 

\bottomrule
\end{tabular}
\end{table}

\section{Conclusion}
\label{sec:conclusion}
In this paper, we introduced ResQu a novel super-resolution framework that integrates quaternion wavelet representations with latent diffusion models, leveraging our encoder to enhance multi-scale conditioning. Our approach establishes a new benchmark in image super-resolution by effectively preserving fine-grained details while maintaining high perceptual quality. Through extensive experiments on diverse datasets, our method consistently outperforms existing techniques across multiple quantitative and qualitative metrics. Notably, our framework demonstrates robustness in handling complex textures and real-world degradation patterns, a key challenge in image super-resolution.
Furthermore, we conducted ablation studies and cross-dataset analyses, to assess the generalization ability of our method. The results highlight the effectiveness of quaternion wavelet embeddings in capturing both structural and textural information.
Overall, this work contributes to advancing super-resolution research by introducing a novel architectural design that integrates quaternion wavelet features, generative priors, and diffusion models. 

\bibliography{references}

@inproceedings{dai2019second,
  title={Second-order attention network for single image super-resolution},
  author={Dai, Tao and Cai, Jianrui and Zhang, Yongbing and Xia, Shu-Tao and Zhang, Lei},
  booktitle={Proceedings of the IEEE/CVF conference on computer vision and pattern recognition},
  pages={11065--11074},
  year={2019}
}

@inproceedings{zhang2021designing,
  title={Designing a practical degradation model for deep blind image super-resolution},
  author={Zhang, Kai and Liang, Jingyun and Van Gool, Luc and Timofte, Radu},
  booktitle={Proceedings of the IEEE/CVF International Conference on Computer Vision},
  pages={4791--4800},
  year={2021}
}

@inproceedings{ji2020real,
  title={Real-world super-resolution via kernel estimation and noise injection},
  author={Ji, Xiaozhong and Cao, Yun and Tai, Ying and Wang, Chengjie and Li, Jilin and Huang, Feiyue},
  booktitle={proceedings of the IEEE/CVF conference on computer vision and pattern recognition workshops},
  pages={466--467},
  year={2020}
}

@inproceedings{wang2018recovering,
  title={Recovering realistic texture in image super-resolution by deep spatial feature transform},
  author={Wang, Xintao and Yu, Ke and Dong, Chao and Loy, Chen Change},
  booktitle={Proceedings of the IEEE conference on computer vision and pattern recognition},
  pages={606--615},
  year={2018}
}

@inproceedings{timofte2017ntire,
  title={Ntire challenge on single image super-resolution: Methods and results},
  author={Timofte, Radu and Agustsson, Eirikur and Van Gool, Luc and Yang, Ming-Hsuan and Zhang, Lei},
  booktitle={Proceedings of the IEEE conference on computer vision and pattern recognition workshops},
  year={2017}
}

@inproceedings{dong2014learning,
  title={Learning a deep convolutional network for image super-resolution},
  author={Dong, Chao and Loy, Chen Change and He, Kaiming and Tang, Xiaoou},
  booktitle={Computer Vision--ECCV 2014: 13th European Conference, Zurich, Switzerland, September 6-12, 2014, Proceedings, Part IV 13},
  pages={184--199},
  year={2014},
  organization={Springer}
}

@inproceedings{agustsson2017ntire,
  title={Ntire challenge on single image super-resolution: Dataset and study},
  author={Agustsson, Eirikur and Timofte, Radu},
  booktitle={Proceedings of the IEEE conference on computer vision and pattern recognition workshops},
  year={2017}
}

@inproceedings{dong2016accelerating,
  title={Accelerating the super-resolution convolutional neural network},
  author={Dong, Chao and Loy, Chen Change and Tang, Xiaoou},
  booktitle={Computer Vision--ECCV 2016: 14th European Conference, Amsterdam, The Netherlands, October 11-14, 2016, Proceedings, Part II 14},
  pages={391--407},
  year={2016},
  organization={Springer}
}

@article{dong2015image,
  title={Image super-resolution using deep convolutional networks},
  author={Dong, Chao and Loy, Chen Change and He, Kaiming and Tang, Xiaoou},
  journal={IEEE transactions on pattern analysis and machine intelligence},
  volume={38},
  number={2},
  pages={295--307},
  year={2015},
  publisher={IEEE}
}

@InBook{aloisi2024waveletdiffusionganimage,
author="Aloisi, Lorenzo
and Sigillo, Luigi
and Uncini, Aurelio
and Comminiello, Danilo",
editor="Esposito, Anna
and Faundez-Zanuy, Marcos
and Morabito, Francesco Carlo
and Pasero, Eros
and Cordasco, Gennaro",
title="A Wavelet Diffusion GAN for Image Super-Resolution",
bookTitle="Neural Networks: Overview of Current Theories and Applications",
year="2026",
publisher="Springer Nature Singapore",
address="Singapore",
pages="425--435",
isbn="978-981-95-4072-3",
doi="10.1007/978-981-95-4072-3_36",
url="https://doi.org/10.1007/978-981-95-4072-3_36"
}

@INPROCEEDINGS{10650042,
  author={Sigillo, Luigi and Gramaccioni, Riccardo Fosco and Nicolosi, Alessandro and Comminiello, Danilo},
  booktitle={International Joint Conference on Neural Networks (IJCNN)}, 
  title={Ship in Sight: Diffusion Models for Ship-Image Super Resolution}, 
  year={2024},
  volume={},
  number={},
  pages={1-8},
  doi={10.1109/IJCNN60899.2024.10650042}}

@inproceedings{
sigilloMWT_2025,
title={Metadata, Wavelet, and Time Aware Diffusion Models for Satellite Image Super Resolution},
author={Luigi Sigillo and Renato Giamba and Danilo Comminiello},
booktitle={ICLR 2025 Workshop on Machine Learning for Remote Sensing (ML4RS)},
year={2025},
url={https://ml-for-rs.github.io/iclr2025/camera_ready/papers/19.pdf}
}

@article{sigillo2024generalizingmedicalimagerepresentations,
title = {Generalizing medical image representations via quaternion wavelet networks},
journal = {Neurocomputing},
volume = {638},
pages = {130195},
year = {2025},
issn = {0925-2312},
doi = {https://doi.org/10.1016/j.neucom.2025.130195},
url = {https://www.sciencedirect.com/science/article/pii/S0925231225008677},
author = {Luigi Sigillo and Eleonora Grassucci and Aurelio Uncini and Danilo Comminiello},
}

@inproceedings{phung2023wavelet,
  title={Wavelet diffusion models are fast and scalable image generators},
  author={Phung, Hao and Dao, Quan and Tran, Anh},
  booktitle={Proceedings of the IEEE/CVF conference on computer vision and pattern recognition},
  pages={10199--10208},
  year={2023}
}

@INPROCEEDINGS{10651227,
  author={Moser, Brian B. and Frolov, Stanislav and Raue, Federico and Palacio, Sebastian and Dengel, Andreas},
  booktitle={2024 International Joint Conference on Neural Networks (IJCNN)}, 
  title={Waving Goodbye to Low-Res: A Diffusion-Wavelet Approach for Image Super-Resolution}, 
  year={2024},
  volume={},
  number={},
  pages={1-8},
  doi={10.1109/IJCNN60899.2024.10651227}}

@INPROCEEDINGS{wang2021real,
  author={Wang, Xintao and Xie, Liangbin and Dong, Chao and Shan, Ying},
  booktitle={IEEE/CVF Int. Conf. on Computer Vision Workshops (ICCVW)}, 
  title={Real-{ESRGAN}: Training Real-World Blind Super-Resolution with Pure Synthetic Data}, 
  year={2021},
 doi={10.1109/ICCVW54120.2021.00217}
}

@inproceedings{chen2022real,
  title={Real-world blind super-resolution via feature matching with implicit high-resolution priors},
  author={Chen, Chaofeng and Shi, Xinyu and Qin, Yipeng and Li, Xiaoming and Han, Xiaoguang and Yang, Tao and Guo, Shihui},
  booktitle={Proceedings of the 30th ACM International Conference on Multimedia},
  pages={1329--1338},
  year={2022}
}

@INPROCEEDINGS{8578168,
  author={Wang, Xintao and Yu, Ke and Dong, Chao and Change Loy, Chen},
  booktitle={IEEE/CVF Conf. on Computer Vision and Pattern Recognition}, 
  title={Recovering Realistic Texture in Image Super-Resolution by Deep Spatial Feature Transform}, 
  year={2018},
  doi={10.1109/CVPR.2018.00070}}

@inproceedings{Wang2018ESRGANES,
  title={{ESRGAN}: Enhanced super-resolution generative adversarial networks},
  author={Wang, Xintao and Yu, Ke and Wu, Shixiang and Gu, Jinjin and Liu, Yihao and Dong, Chao and Qiao, Yu and Change Loy, Chen},
  booktitle={Proceedings of the European Conf. on computer vision (ECCV) workshops},
  pages={0--0},
  year={2018}
}

@article{Menon2020PULSESP,
  title={PULSE: Self-Supervised Photo Upsampling via Latent Space Exploration of Generative Models},
  author={Sachit Menon and Alexandru Damian and Shijia Hu and Nikhil Ravi and Cynthia Rudin},
  journal={IEEE/CVF Conf. on Computer Vision and Pattern Recognition (CVPR)},
  year={2020},
  pages={2434-2442}
}

@article{Rombach2021HighResolutionIS,
  title={High-Resolution Image Synthesis with Latent Diffusion Models},
  author={Robin Rombach and A. Blattmann and Dominik Lorenz and Patrick Esser and Bj{\"o}rn Ommer},
  journal={IEEE/CVF Conf. on Computer Vision and Pattern Recognition (CVPR)},
  year={2021},
}

@article{ledig2017photorealistic,
  title={Photo-Realistic Single Image Super-Resolution Using a Generative Adversarial Network},
  author={Christian Ledig and Lucas Theis and Ferenc Husz{\'a}r and Jose Caballero and Andrew P. Aitken and Alykhan Tejani and Johannes Totz and Zehan Wang and Wenzhe Shi},
  journal={IEEE Conf. on Computer Vision and Pattern Recognition (CVPR)},
  year={2016},
  pages={105-114},
  
}

@INPROCEEDINGS{9879163,
  author={Choi, Jooyoung and Lee, Jungbeom and Shin, Chaehun and Kim, Sungwon and Kim, Hyunwoo and Yoon, Sungroh},
  booktitle={IEEE/CVF Conf. on Computer Vision and Pattern Recognition (CVPR)}, 
  title={Perception Prioritized Training of Diffusion Models}, 
  year={2022},
  volume={},
  number={},
  doi={10.1109/CVPR52688.2022.01118}}

@article{yue2024resshift,
  title={Resshift: Efficient diffusion model for image super-resolution by residual shifting},
  author={Yue, Zongsheng and Wang, Jianyi and Loy, Chen Change},
  journal={Advances in Neural Information Processing Systems},
  volume={36},
  year={2024}
}

@InProceedings{Ruiz_2023_CVPR,
    author    = {Ruiz, Nataniel and Li, Yuanzhen and Jampani, Varun and Pritch, Yael and Rubinstein, Michael and Aberman, Kfir},
    title     = {DreamBooth: Fine Tuning Text-to-Image Diffusion Models for Subject-Driven Generation},
    booktitle = {Proceedings of the IEEE/CVF Conference on Computer Vision and Pattern Recognition (CVPR)},
    month     = {June},
    year      = {2023},
    pages     = {22500-22510}
}

@article{huang2024wavedm,
  title={Wavedm: Wavelet-based diffusion models for image restoration},
  author={Huang, Yi and Huang, Jiancheng and Liu, Jianzhuang and Yan, Mingfu and Dong, Yu and Lyu, Jiaxi and Chen, Chaoqi and Chen, Shifeng},
  journal={IEEE Transactions on Multimedia},
  year={2024},
  publisher={IEEE}
}

@inproceedings{wu2024seesr,
  title={Seesr: Towards semantics-aware real-world image super-resolution},
  author={Wu, Rongyuan and Yang, Tao and Sun, Lingchen and Zhang, Zhengqiang and Li, Shuai and Zhang, Lei},
  booktitle={Proceedings of the IEEE/CVF conference on computer vision and pattern recognition},
  pages={25456--25467},
  year={2024}
}

@inproceedings{gao2023implicit,
  title={Implicit diffusion models for continuous super-resolution},
  author={Gao, Sicheng and Liu, Xuhui and Zeng, Bohan and Xu, Sheng and Li, Yanjing and Luo, Xiaoyan and Liu, Jianzhuang and Zhen, Xiantong and Zhang, Baochang},
  booktitle={Proceedings of the IEEE/CVF conference on computer vision and pattern recognition},
  pages={10021--10030},
  year={2023}
}

@inproceedings{shang2024resdiff,
  title={Resdiff: Combining cnn and diffusion model for image super-resolution},
  author={Shang, Shuyao and Shan, Zhengyang and Liu, Guangxing and Wang, LunQian and Wang, XingHua and Zhang, Zekai and Zhang, Jinglin},
  booktitle={Proceedings of the AAAI Conference on Artificial Intelligence},
  volume={38},
  number={8},
  pages={8975--8983},
  year={2024}
}

@article{wang2024lavie,
  title={Lavie: High-quality video generation with cascaded latent diffusion models},
  author={Wang, Yaohui and Chen, Xinyuan and Ma, Xin and Zhou, Shangchen and Huang, Ziqi and Wang, Yi and Yang, Ceyuan and He, Yinan and Yu, Jiashuo and Yang, Peiqing and others},
  journal={International Journal of Computer Vision},
  pages={1--20},
  year={2024},
  publisher={Springer}
}

@InProceedings{Takagi_2023_CVPR,
    author    = {Takagi, Yu and Nishimoto, Shinji},
    title     = {High-Resolution Image Reconstruction With Latent Diffusion Models From Human Brain Activity},
    booktitle = {Proceedings of the IEEE/CVF Conference on Computer Vision and Pattern Recognition (CVPR)},
    month     = {June},
    year      = {2023},
    pages     = {14453-14463}
}

@ARTICLE{10183382,
  author={Hsu, Wei-Yen and Jian, Pei-Wen},
  journal={IEEE Transactions on Neural Networks and Learning Systems}, 
  title={Wavelet Pyramid Recurrent Structure-Preserving Attention Network for Single Image Super-Resolution}, 
  year={2024},
  volume={35},
  number={11},
  pages={15772-15786},
  doi={10.1109/TNNLS.2023.3289958}}

@INPROCEEDINGS{lopez2025guessithinkstreamlined,
  author={Lopez, Eleonora and Sigillo, Luigi and Colonnese, Federica and Panella, Massimo and Comminiello, Danilo},
  booktitle={ICASSP 2025 - IEEE International Conference on Acoustics, Speech and Signal Processing (ICASSP)}, 
  title={Guess What I Think: Streamlined {EEG}-to-Image Generation with Latent Diffusion Models}, 
  year={2025},
  volume={},
  number={},
  pages={1-5},
  doi={10.1109/ICASSP49660.2025.10890059}}

@inproceedings{kim2024beyond,
  title={Beyond Image Super-Resolution for Image Recognition with Task-Driven Perceptual Loss},
  author={Kim, Jaeha and Oh, Junghun and Lee, Kyoung Mu},
  booktitle={Proceedings of the IEEE/CVF Conference on Computer Vision and Pattern Recognition},
  pages={2651--2661},
  year={2024}
}

@article{lepcha2023image,
  title={Image super-resolution: A comprehensive review, recent trends, challenges and applications},
  author={Lepcha, Dawa Chyophel and Goyal, Bhawna and Dogra, Ayush and Goyal, Vishal},
  journal={Information Fusion},
  volume={91},
  pages={230--260},
  year={2023},
  publisher={Elsevier}
}

@inproceedings{lin2025diffbir,
  title={Diffbir: Toward blind image restoration with generative diffusion prior},
  author={Lin, Xinqi and He, Jingwen and Chen, Ziyan and Lyu, Zhaoyang and Dai, Bo and Yu, Fanghua and Qiao, Yu and Ouyang, Wanli and Dong, Chao},
  booktitle={European Conference on Computer Vision},
  pages={430--448},
  year={2025},
  organization={Springer}
}

@article{selesnick2005dual,
  title={The dual-tree complex wavelet transform},
  author={Selesnick, Ivan W and Baraniuk, Richard G and Kingsbury, Nick C},
  journal={IEEE signal processing magazine},
  volume={22},
  number={6},
  pages={123--151},
  year={2005},
  publisher={IEEE}
}

@article{chan2008coherent,
  title={Coherent multiscale image processing using dual-tree quaternion wavelets},
  author={Chan, Wai Lam and Choi, Hyeokho and Baraniuk, Richard G},
  journal={IEEE Transactions on Image Processing},
  volume={17},
  number={7},
  pages={1069--1082},
  year={2008},
  publisher={IEEE}
}

@book{barford1992introduction,
  title={An introduction to wavelets},
  author={Barford, Lee A and Fazzio, R Shane and Smith, David R},
  year={1992},
  publisher={Hewlett Packard}
}

@inproceedings{Zhang2018ImageSU,
  title={Image Super-Resolution Using Very Deep Residual Channel Attention Networks},
  author={Yulun Zhang and Kunpeng Li and Kai Li and Lichen Wang and Bineng Zhong and Yun Raymond Fu},
  booktitle={European Conf. on Computer Vision},
  year={2018}
}

@inproceedings{chan2004quaternion,
  title={Quaternion wavelets for image analysis and processing},
  author={Chan, Wai Lam and Choi, Hyeokho and Baraniuk, Richard},
  booktitle={2004 International Conference on Image Processing, 2004. ICIP'04.},
  volume={5},
  pages={3057--3060},
  year={2004},
  organization={IEEE}
}

@article{wang2023exploiting,
  title={Exploiting diffusion prior for real-world image super-resolution},
  author={Wang, Jianyi and Yue, Zongsheng and Zhou, Shangchen and Chan, Kelvin CK and Loy, Chen Change},
  journal={International Journal of Computer Vision},
  pages={1--21},
  year={2024},
  publisher={Springer}
}

@ARTICLE{saharia2021image,
  author={Saharia, Chitwan and Ho, Jonathan and Chan, William and Salimans, Tim and Fleet, David J. and Norouzi, Mohammad},
  journal={IEEE Trans. on Pattern Analysis and Machine Intelligence}, 
  title={Image Super-Resolution via Iterative Refinement}, 
  year={2023},
  volume={45},
  number={4},
  doi={10.1109/TPAMI.2022.3204461}
}

@inproceedings{
dosovitskiy2021image,
title={An Image is Worth 16x16 Words: Transformers for Image Recognition at Scale},
author={Alexey Dosovitskiy and Lucas Beyer and Alexander Kolesnikov and Dirk Weissenborn and Xiaohua Zhai and Thomas Unterthiner and Mostafa Dehghani and Matthias Minderer and Georg Heigold and Sylvain Gelly and Jakob Uszkoreit and Neil Houlsby},
booktitle={Int. Conf. on Learning Representations},
year={2021},

}

@inproceedings{
brock2019large,
title={Large Scale {GAN} Training for High Fidelity Natural Image Synthesis},
author={Andrew Brock and Jeff Donahue and Karen Simonyan},
booktitle={Int. Conf. on Learning Representations},
year={2019},

}

@inproceedings{wei2020component,
  title={Component divide-and-conquer for real-world image super-resolution},
  author={Wei, Pengxu and Xie, Ziwei and Lu, Hannan and Zhan, Zongyuan and Ye, Qixiang and Zuo, Wangmeng and Lin, Liang},
  booktitle={Computer Vision--ECCV 2020: 16th European Conference, Glasgow, UK, August 23--28, 2020, Proceedings, Part VIII 16},
  pages={101--117},
  year={2020},
  organization={Springer}
}

@inproceedings{cai2019toward,
  title={Toward real-world single image super-resolution: A new benchmark and a new model},
  author={Cai, Jianrui and Zeng, Hui and Yong, Hongwei and Cao, Zisheng and Zhang, Lei},
  booktitle={Proceedings of the IEEE/CVF international conference on computer vision},
  pages={3086--3095},
  year={2019}
}

@article{vetterli1992wavelets,
  title={Wavelets and filter banks: Theory and design},
  author={Vetterli, Martin and Herley, Cormac},
  journal={IEEE transactions on signal processing},
  year={1992}
}

@ARTICLE{4526699,
  author={Chan, Wai Lam and Choi, Hyeokho and Baraniuk, Richard G.},
  journal={IEEE Transactions on Image Processing}, 
  title={Coherent Multiscale Image Processing Using Dual-Tree Quaternion Wavelets}, 
  year={2008},
  volume={17},
  number={7},
  pages={1069-1082},
  doi={10.1109/TIP.2008.924282}}

@article{zhancheng2020medical,
  title={Medical image fusion based on quaternion wavelet transform},
  author={Zhancheng, Zhang and Xiaoqing, Luo and Mengyu, Xiong and Zhiwen, Wang and Kai, Li},
  journal={Journal of Algorithms \& Computational Technology},
  volume={14},
  year={2020},
  publisher={SAGE Publications Sage UK: London, England}
}

@INPROCEEDINGS{Chan2004QWT,
  author={Wai Lam Chan and Hyeokho Choi and Baraniuk, R.},
  booktitle={2004 International Conference on Image Processing, 2004. ICIP '04.}, 
  title={Quaternion wavelets for image analysis and processing}, 
  year={2004},
  volume={5},
  number={},
  pages={3057-3060 Vol. 5},
  doi={10.1109/ICIP.2004.1421758}}

@ARTICLE{grouse,
  author={Grassucci, Eleonora and Sigillo, Luigi and Uncini, Aurelio and Comminiello, Danilo},
  journal={IEEE Signal Processing Letters}, 
  title={{GROUSE}: A Task and Model Agnostic Wavelet- Driven Framework for Medical Imaging}, 
  year={2023},
  volume={30},
  number={},
  pages={1397-1401},
  doi={10.1109/LSP.2023.3321554}}

@INPROCEEDINGS{swinir,
  author={Liang, Jingyun and Cao, Jiezhang and Sun, Guolei and Zhang, Kai and Van Gool, Luc and Timofte, Radu},
  booktitle={2021 IEEE/CVF Int. Conf. on Computer Vision Workshops (ICCVW)}, 
  title={Swin{IR}: Image Restoration Using Swin Transformer}, 
  year={2021},
  doi={10.1109/ICCVW54120.2021.00210}}

@article{ssim,
author = {Wang, Zhou and Bovik, Alan and Sheikh, Hamid and Simoncelli, Eero},
year = {2004},
month = {05},
pages = {600 - 612},
title = {Image Quality Assessment: From Error Visibility to Structural Similarity},
volume = {13},
journal = {Image Processing, IEEE Trans. on},
doi = {10.1109/TIP.2003.819861}
}

@inproceedings{
Song2020DenoisingDI,
title={Denoising Diffusion Implicit Models},
author={Jiaming Song and Chenlin Meng and Stefano Ermon},
booktitle={Int. Conf. on Learning Representations},
year={2021},

}
\bibliographystyle{IEEEtran}

\end{document}